# A Metric for Evaluating Neural Input Representation in Supervised Learning Networks

Richard R. Carrillo[1,2]*, Francisco Naveros[1,2], Eduardo Ros[1,2] and Niceto R. Luque[1,2,3]

[1] Department of Computer Architecture and Technology, University of Granada, Granada, Spain, [2] Centro de Investigación en Tecnologías de la Información y de las Comunicaciones (CITIC-UGR), University of Granada, Granada, Spain, [3] Aging in Vision and Action, Institut de la Vision, Inserm–UPMC–CNRS, Paris, France

Supervised learning has long been attributed to several feed-forward neural circuits within the brain, with particular attention being paid to the cerebellar granular layer. The focus of this study is to evaluate the input activity representation of these feed-forward neural networks. The activity of cerebellar granule cells is conveyed by parallel fibers and translated into Purkinje cell activity, which constitutes the sole output of the cerebellar cortex. The learning process at this parallel-fiber-to-Purkinje-cell connection makes each Purkinje cell sensitive to a set of specific cerebellar states, which are roughly determined by the granule-cell activity during a certain time window. A Purkinje cell becomes sensitive to each neural input state and, consequently, the network operates as a function able to generate a desired output for each provided input by means of supervised learning. However, not all sets of Purkinje cell responses can be assigned to any set of input states due to the network's own limitations (inherent to the network neurobiological substrate), that is, not all input-output mapping can be learned. A key limiting factor is the representation of the input states through granule-cell activity. The quality of this representation (e.g., in terms of heterogeneity) will determine the capacity of the network to learn a varied set of outputs. Assessing the quality of this representation is interesting when developing and studying models of these networks to identify those neuron or network characteristics that enhance this representation. In this study we present an algorithm for evaluating quantitatively the level of compatibility/interference amongst a set of given cerebellar states according to their representation (granule-cell activation patterns) without the need for actually conducting simulations and network training. The algorithm input consists of a real-number matrix that codifies the activity level of every considered granule-cell in each state. The capability of this representation to generate a varied set of outputs is evaluated geometrically, thus resulting in a real number that assesses the goodness of the representation.

## INTRODUCTION

Activity-dependent synaptic plasticity is at the core of the neural mechanisms underlying learning (Elgersma and Silva, 1999; Abbott and Regehr, 2004) i.e., several forms of spike-timing-dependent plasticity (STDP) are reported to mediate long-lasting modifications of synapse efficacy in the brain (Markram et al., 1997; Song et al., 2000; Dan and Poo, 2004). These mechanisms regulate the activity generated by a neuron in response to its current synaptic activity (Kempter et al., 1999; Cooke and Bliss, 2006), thus adjusting this generated neural activity for certain input patterns (input activity). This readout neuron, therefore, learns to "recognize" and react to these patterns with a corresponding intensity (output activity). This process provides the basis for supervised learning, in which a "teaching" activity drives the synaptic efficacy changes (**Figure 1**) seeking an adequate input-to-output-activity transformation function (Knudsen, 1994; Doya, 2000).

***The cerebellum*** (for vestibulo-ocular reflex and motor control in vertebrates) (Miles and Eighmy, 1980; Kawato et al., 2011), and the ***inferior colliculus***[1] (for audio-visual alignment of the barn owl) (Knudsen, 2002; Takahashi, 2010) are just two examples of neural circuits to which supervised learning is attributed (See Knudsen, 1994 for a review of supervised learning examples in the brain).

The input-to-output transformation of these supervised-learning circuits associates one desired output value (e.g., encoded by firing rate) to a certain input state from the set of considered input states. This set of spatiotemporal states is encoded by a population of granule cells in the cerebellum (Yamazaki and Tanaka, 2007), and it is generated from multimodal information conveyed by mossy fibers (Sawtell, 2010). This input-information transformation by the granule cells has long been believed to expand the coding space, thus enhancing the capacity of output neurons (i.e., Purkinje cells as readout neurons) to generate desired responses for each state (Marr, 1969; Albus, 1971; Schweighofer et al., 2001; Cayco-Gajic et al., 2017). In the case of the midbrain, the central-nucleus activity of the inferior colliculus encodes auditory information (input state) that generates the desired output responses (map of space) in external-nucleus neurons of the inferior colliculus (Friedel and van Hemmen, 2008). This input-to-output mapping is hypothesized to implement an audio-visual alignment function (Singheiser et al., 2012). It is important to note that the cerebellum and the inferior colliculus are but two examples of neural circuits whose input state representation is suitable for being assessed and studied, amongst other possible neural circuits e.g., networks undergoing reinforcement learning whose input state representation could also be considered.

In this work, we aim to evaluate the input-to-output mapping capabilities of an output neuron, that is, its capacity to recognize different input states and generate a different output for each state. Since some sets of output values cannot be associated to certain sets of input states, the output-neuron mapping capabilities are thus constrained. The capacity to generate a diverse set of output values is directly affected by the particular activity codification corresponding to the input states (i.e., input representation), understanding by this activity codification the pattern of synapses that are activated during each input state. To mention some examples, when all the input states are represented by 0 (no activity), the only possible value for all the output states is 0 (worst-case representation), whereas when each state is represented by activity in one or more synapses that are not shared by other states, any output value can be associated to any state (best-case representation). Between these two extreme representations, there are a varied set of possibilities that include more realistic representations.

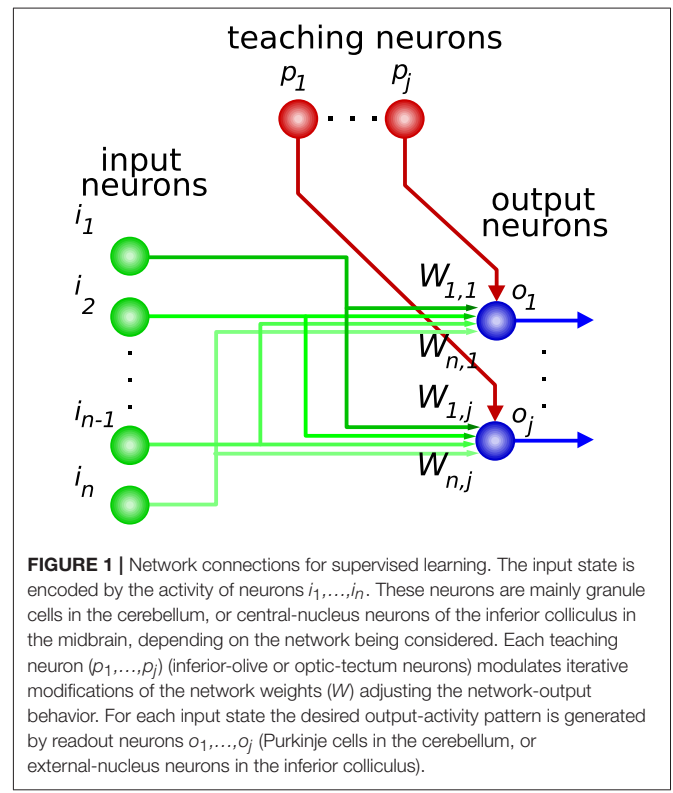

**FIGURE 1 |** Network connections for supervised learning. The input state is encoded by the activity of neurons $i_1, ..., i_n$. These neurons are mainly granule cells in the cerebellum, or central-nucleus neurons of the inferior colliculus in the midbrain, depending on the network being considered. Each teaching neuron ($p_1, ..., p_j$) (inferior-olive or optic-tectum neurons) modulates iterative modifications of the network weights ($W$) adjusting the network-output behavior. For each input state the desired output-activity pattern is generated by readout neurons $o_1, ..., o_j$ (Purkinje cells in the cerebellum, or external-nucleus neurons in the inferior colliculus).

For this purpose, we present an evaluation function and its corresponding algorithmic definition able to analytically assess the fitness of a particular representation of a set of input states. The function focuses on evaluating a single output neuron, since all output neurons receiving the same inputs have the same learning capacity. This evaluation would be the equivalent to conducting an infinite number of network simulations and to calculating the average error performed when learning every possible output value. It is worth mentioning that calculating the best network weights to approximate a given output is equivalent to solving a (non-negative)-least-squares problem (Lawson and Hanson, 1974). However, in a least-squares problem, the set of input states and the corresponding output value for each state are specified, and the algorithm obtains the best set of network weights. In the algorithm that we propose, the input-state representation constitutes the sole algorithm input, and its output is a real number indicating the suitability of the set of input states to generate any output.

---

[1] We use the mammal-anatomy term inferior colliculus instead of the avian mesencephalicus lateralis dorsolis for homogeneity with previous literature.

## Limitations of the Evaluation Function

Neural-circuit modeling has proven itself as a fast and versatile method to study and understand the operation of specific neural circuits. Consequently, a large number of models with increasing levels of complexity and degree of detail are being developed (McDougal et al., 2017), yet neural modeling usually requires some assumptions to be made. These assumptions are usually made based on the operation of the modeled circuits. This is particularly true in developing functional models that must solve a task or mimic an expected behavior. Some of these assumptions determine the efficiency and flexibility of the input-output mapping of a supervised-learning circuit (Luque et al., 2011a). A function for evaluating the input representation must assess this representation according to the model input-output mapping capabilities. Our evaluation function is tailored according to those models of supervised-learning circuits that assume that (i) synaptic weights converge to an optimal solution (as suggested in Wong and Sideris, 1992; Dean and Porrill, 2014), (ii) the effect of simultaneous input activity on several readout-neuron synapses is additive, and (iii) the input state is codified through excitatory activity.

When the effect of simultaneous synaptic activity is assumed to be additive, the output neuron can be approximated by a weighted sum of inputs. This is a common approach considered in functional models (Albus, 1975; Kawato and Gomi, 1992; Tyrrell and Willshaw, 1992; Rucci et al., 1997; Raymond and Lisberger, 1998; Doya, 1999; Kardar and Zee, 2002; Friedel and van Hemmen, 2008; Garrido et al., 2013; Clopath et al., 2014). A readout neuron is then not able to completely differentiate and decouple inputs that are linearly dependent (Haykin, 1999) and our evaluation function, therefore, penalizes those representations that include many linearly-dependent states.

In regard to the codification of input states, these functional models usually fall in one of two main groups: (i) those that equally use excitatory and inhibitory activity (for example, representing the input activity through values that can be positive or negative, or allowing negative synaptic weights) and (ii) those that only use excitatory activity to codify states. It is worth mentioning that probably both groups entail simplification:

The first group assumes that excitation and inhibition play equivalent and opposite roles in the state codification (Fujita, 1982; Wong and Sideris, 1992; minimal model of Clopath et al., 2014). However, there exists a great imbalance between the number of excitatory and inhibitory neurons in the aforementioned networks (cerebellum and inferior colliculus). In the rat cerebellar cortex these inhibitory input neurons (GABAergic molecular layer interneurons), which converge upon the Purkinje cells, are only a small fraction (about 2.3%) of the granule-cell number (excitatory neurons) (Korbo et al., 1993). In the external nucleus of the inferior colliculus of the barn owl only about 15% of the neurons are GABAergic (Carr et al., 1989). The second group of models does not include inhibitory input in the state codification (Carrillo et al., 2008a; Friedel and van Hemmen, 2008; Luque et al., 2011a; Garrido et al., 2013). However, in the cerebellum and inferior colliculus, the output neurons receive inhibitory inputs (Häusser and Clark, 1997; Knudsen, 2002).

Several metrics have been used in previous works to analytically evaluate the input-to-output mapping capabilities of models belonging to the first group: e.g., a calculation based on the eigenvalues of the covariance matrix of the input activity matrix (Cayco-Gajic et al., 2017) and the rank of the input activity matrix (Barak et al., 2013). From the best of our knowledge, no equivalent analytical calculations have been previously proposed for models belonging to the second group, therefore, the evaluation function that we present is conceived to analytically asses the input-state representation of models in this second group.

Note that the assumption of an excitatory-activity codification for the input states (second group) constrains the neural-network model to zero-or-positive inputs and synaptic weights ($w_{i,j} \geq 0$). This premise results in an increment of the proposed-algorithm complexity in comparison to evaluation functions for models that do not constrain the input sign (first group).

## Applications of the Evaluation Function

Our algorithm is meant to be exploited when developing and studying models of supervised-learning networks. Even when these network models derive from biological data (Solinas et al., 2010; Garrido et al., 2016) some free parameters must usually be estimated or tuned to reproduce results from biological experimentation (Carrillo et al., 2007, 2008b; Masoli et al., 2017) or to render the model functional for solving a specific task (Luque et al., 2011b). In particular, the free parameters of the circuits that generate the input for a supervised-learning network can be adjusted according to the quality of this generated input-state representation. Likewise, some characteristics of these state-generating circuits are usually key for the performance of their processing (such as their connectivity). These characteristics can be identified by means of the quality of the generated state representation (Cayco-Gajic et al., 2017). When this optimization of the model parameters is performed automatically (e.g., by a genetic algorithm) (Lynch and Houghton, 2015; Martínez-Cañada et al., 2016; Van Geit et al., 2016) the proposed evaluation function could be directly used as the cost function for guiding the search algorithm. Apart from tuning intrinsic network parameters, the input-state representation can also be considered for improvement when the network input is reproduced and refined (Luque et al., 2012), since a complete characterization of the network input activity is usually not tractable (Feng, 2003).

## MATERIALS AND METHODS

### Representation of the Neural Activity

In order for the proposed algorithm to evaluate the representation of input states we need to codify this input (or output) information numerically. Most information in nervous circuits is transmitted by action potentials (spikes), i.e., at the time when these spikes occur. It is general practice to divide time into slots and translate the neural activity (spike times) within each time slot into an activity effectiveness number.

This number is then used to encode the neural activity capacity to excite a target neuron, that is, to depolarize its membrane potential. The cyclic presentation of inputs in time windows or slots is compatible with the cerebellar theory about Purkinje-cell signal sampling driven by background activity oscillations (D'Angelo et al., 2009; Gandolfi et al., 2013). Similarly, the activity in the inferior colliculus is intrinsically coupled with other anatomically connected areas through particular frequency bands (Stitt et al., 2015). A common straightforward translation consists of counting the number of spikes in each slot (Gerstner and Kistler, 2002). This translation represents an input state as an $n$-element vector, where $n$ denotes the number of input neurons. We assume that all input states have equal occurrence probability and relevance. This assumption reduces the input of our evaluation algorithm to an $m$-by-$n$ matrix of integer numbers, which represents the set of distinct input states. We will denote this matrix as $C$, where $m$ represents the number of different input states. Element $c_{k,l}$ corresponds to the activity effectiveness of the $l$-th input neuron for the $k$-th input state. Each input state requires one time slot to be presented (see **Figure 2**). More meticulous translations may also consider the spike dispersion within the time slot. This latter translation would imply using real numbers instead of integers to represent the matrix $C$, therefore we generalize the proposed input-evaluation algorithm to support a $C$ real matrix in order to provide compatibility with this potential representation.

The readout-neuron output for the $k$-th distinct input state in $C$ is defined by:

$$d_k = \sum_{l=1}^{n} w_l c_{k,l} \tag{1}$$

where $w_l$ is the weight of the $l$-th connection, $d_k$ is the output corresponding to the $k$-th input state and $n$ is the number of input neurons. This expression is equivalent to the matrix equation:

$$d = Cw \tag{2}$$

where $d$ and $w$ are column vectors containing the output for all the distinct input states and the synaptic weights, respectively. That is to say, vector $d$ contains $m$ components (the readout-neuron output for the $m$ input states) and vector $w$ contains $n$ components (the readout-neuron synaptic weight corresponding to the $n$ input neurons). We assume an additive effect of the inputs, positive input values and positive weights (see Limitations of the evaluation function for additional information).

## Definition of the Evaluation Function

For a given representation of input states (matrix $C$) and a hypothetical desired output for each state (vector $d_{des}$), the calculated weights (vector $\hat{w}$) would generate an approximate output (vector $\hat{d}$, with one element per input state). The inaccuracy (error) of $\hat{d}$ can be measured by means of the residual vector, which is defined as the difference between vector $d_{des}$ and vector $\hat{d}$. This residual vector can be reduced to a single number

by summing its squared elements, obtaining the squared $l^2$ norm of the residual vector:

$$\left\| d_{des} - \hat{a} \right\|_2^2 = \left\| d_{des} - C\hat{w} \right\|_2^2 \tag{3}$$

where $\|v\|_2$ denotes the $l^2$ norm of vector $v$. This error measurement corresponds to the squared Euclidean distance between the vector of actual outputs of the readout neuron and the vector of desired outputs (each vector component codifies the output for a different input state). This distance equally weights the error committed by the readout neuron for all the input states. The calculation of the presented evaluation function is based on this distance, therefore the function assumes that all input states have the same relevance (and occurrence probability).

Assuming that the learning mechanism is able to achieve an optimal output, $\hat{w}$ represents the weight vector that minimizes the error:

$$\hat{w} = \arg \min_w \left\| d - Cw \right\|_2^2 \ subject \ to \ w \geq 0 \tag{4}$$

akin to the problem of non-negative least squares (Chen and Plemmons, 2009).

It is obvious that the suitability of a matrix $C$ to make the readout neuron generate a particular vector of outputs ($d$) depends on the concrete value of the desired outputs ($d_{des}$); nonetheless, we aim to measure the goodness of $C$ by itself, without considering one particular desired output. Matrix $C$ is, thus, evaluated for every possible desired readout-neuron output and every specified input state. To this end, our algorithm calculates the average of the committed output error (measured by the squared $l^2$ norm of the residual vector) for every possible vector of desired outputs. We define this set of all the considered output vectors (possible values of $d_{des}$) as $S$. Each of these output vectors contains $m$ components (one output value per input state), and thus, $S$ can be regarded as an $m$-dimensional space (of positive real numbers, since we are assuming positive outputs). The coordinates of each point of this space (vector $d_{des}$) represent the $m$ neuron output values for the $m$ input states. Since $S$ is an infinite set, the averaging becomes an integral divided by a volume. Thus, the error calculation for a matrix $C$ ($Ir(C)$) becomes:

$$Ir(C) = \frac{\int_S \left\| s - C\hat{w}_s \right\|_2^2 \, ds}{Vol(S)} \tag{5}$$

where $Vol(S)$ denotes the volume of the set S and $\hat{w}_s$ represents the changing weight vector that minimizes the error depending on $s$ (and hence depending on $C$) (Equation 4). This calculation can be regarded as the sum of the error for each vector $d_{des}$ in the set, divided by the number of vectors in the set. We propose this function ($Ir(C)$) as a measurement of the quality of the input-state representation of $C$. This calculation, however, is simplified to enable its implementation as described below.

Note that if the readout neuron is able to achieve a specific $d$ output value with a particular input matrix $C$, it can also obtain an $\alpha d$ value (being $\alpha$ a positive scalar) with the same input by multiplying $w$ by $\alpha$ (see Equation 2). Consequently, calculating the error for a bounded interval of desired outputs is enough to consider all possible outputs. For the sake of simplicity, we consider the interval $[0, 1]$ of desired readout-neuron outputs for any input state. Since a matrix $C$ specifies $m$ input states, the set of possible desired outputs ($S$) comprises all the vector values in the multi-dimensional interval $[0, 1]^m$, being $m$ the number of rows of $C$. The volume of this set ($Vol(S)$) is 1, simplifying the calculation. That is to say, in the case $m = 2$ the set $S$ comprises all the points in a square of size 1 (and area 1). We denote this set of points by $[0, 1]^2$ in a two-dimensional space (see **Figure 3**). The 2 coordinates of each point comprise the desired output (elements of vector $d_{des}$) for the 2 input states. In the case $m = 3$ the set $S$ comprises all the points in a cube ($[0, 1]^3$, see **Figure 4**) in a three-dimensional space, and in the general case $m$ the set $S$ comprises all the points in a hypercube ($[0, 1]^m$) in an $m$-dimensional space.

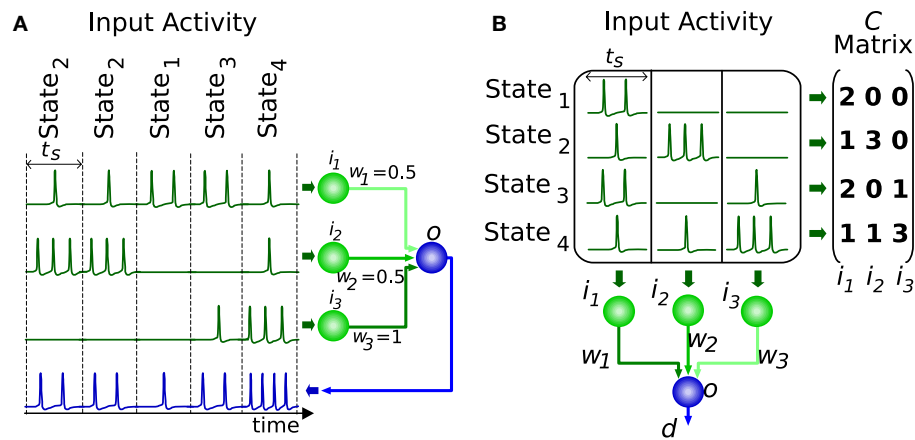

**FIGURE 2 |** Representation of input states. This figure illustrates a network with 3 input neurons ($i_1$, $i_2$, $i_3$) and 4 different input states. **(A)** The input states are presented to the output neuron throughout time. Each state requires one time slot ($t_s$) to be presented to the output neuron ($o$). **(B)** The set of distinct input states is represented by the matrix $C$. Each element of this matrix encodes the number of spikes elicited by the input neuron corresponding to that matrix column. Each matrix row denotes the input activity during a particular state. The output $d$ is computed by the weighted sum ($w_1$, $w_2$, $w_3$) of inputs for each state.

## Geometric Representation of the Input States

By designating the $k$-th column vector of $C$ as $u_k$, Equation (2) can be rewritten as:

$$d = \sum_{k=1}^{n} u_k w_k \qquad (6)$$

Vector $u_k$ contains the $m$ activity values of the $k$-th input neuron corresponding to the $m$ input states. Hence, the value of vector $d$ (which contains the readout-neuron output for all these input states) can be regarded as a linear combination of columns of $C$ with coefficients $w_k$. We consider that these weights can only take positive values ($w_k \geq 0$), therefore the set of all possible outputs (values of vector $d$) is bounded by vectors $u_k$. In the case $m = 3$, when these vectors are represented graphically starting from a common point (origin), they form the edges of a pyramid with apex in the origin and an infinite height (see the empty space of the cube in **Figure 4**). The points of space (vectors) inside this pyramid constitute the set of all the values of $d$ that the readout neuron can generate; this set is a cone called the (convex) conical hull of C ($coni(C)$) (Hiriart-Urruty and Lemaréchal, 1993). This weighted combination of several vectors $u_k$ to obtain a value of $d$ (Equation 6) is called the (convex) conical combination (Jeter, 1986). We are, thus, considering that the output neuron is performing a conical combination of the inputs.

Some of the vectors $u_k$ of a matrix $C$ can be obtained by conical combination of other vectors $u_k$ of the same matrix, and thus, they add redundant information to the conical-hull definition. Therefore, the set of all possible readout-neuron outputs for a given matrix $C$ ($coni(C)$) can be obtained with just the subset of $C$ columns (that is, input neurons) required to define the same conical hull. These subset column vectors are known as the cone's extreme rays (Avis, 1980).

This reduction of redundant input neurons can be illustrated with a 4-input-neuron network and the following 2-state representation:

$$C = \begin{pmatrix} 1 & 3 & 1 & 2 \\ 1 & 2 & 0 & 1 \end{pmatrix}$$

Column vectors of $C$ can be represented in a two-dimensional space (see **Figure 3**). Only vectors $u_1$ and $u_3$ (the first and third input neurons) of this representation are required to define its conical hull (extreme rays), whereas the second and fourth input neurons are just adding redundant information. Therefore, only vectors $u_1$ and $u_3$ need to be considered to evaluate the quality of this representation of input states ($C$).

As previously stated, the error evaluation for this matrix ($Ir(C)$) is obtained by summing the squared $l^2$ norm of the residual vector (output error) for all the desired outputs in the square $[0, 1]^2$. This sum is calculated by a definite integral of this squared $l^2$ norm over $[0, 1]^2$. This squared $l^2$ norm corresponds to the squared Euclidean distance from a point in the square ($d_{des}$) to the closest point ($d$) in the conical hull. If the output $d_{des}$ can actually be generated by the readout neuron, $d_{des}$ is located within the conical hull and this distance becomes 0. Since these 0 values of the distance do not affect the total sum, the integration is only calculated over the space (polygons) of the square that are not covered by the hull. This area ($[0, 1]^2$-coni(C)) is represented by a gray triangle in **Figure 3**. In a two-dimensional space (two input states), up to two polygons can be considered for integration, one on each side of the conical hull.

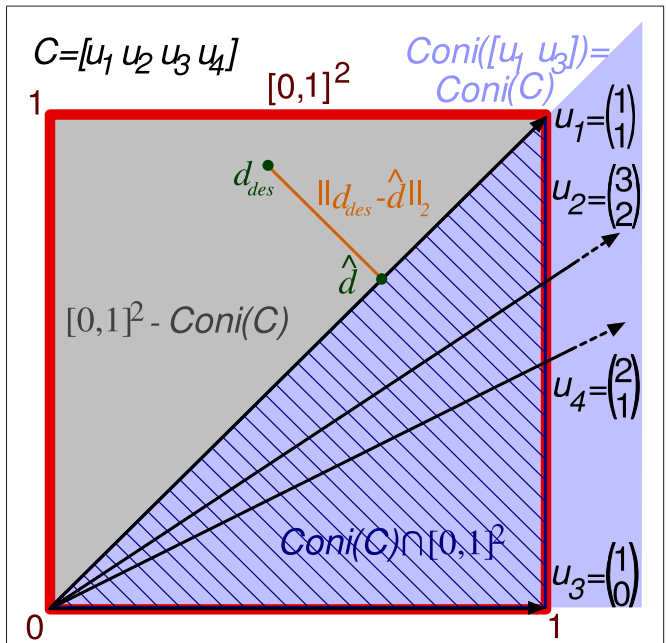

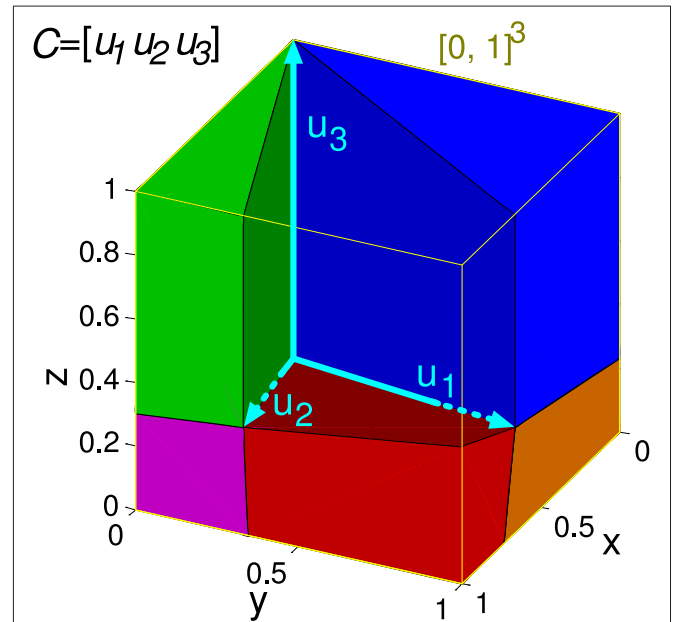

**FIGURE 3 |** Evaluation of the representation error for a 2-state-and-4-input-neuron network. The output error performed by a readout neuron using the 2-input-state representation matrix C is depicted in a two-dimensional space. The square of size 1 (delimited by the red line) contains the space of potentially desired outputs considered by the evaluation function ($[0, 1]^2$). That is, the 2 coordinates of each point in this square codify a potentially desired output for the 2 input states. $u_1$, $u_2$, $u_3$, and $u_4$ refer to the columns of C represented as vectors. Coni(C) is the conical hull of the column vectors of C, represented by the light purple cone area. This cone covers the space of outputs that can be generated by the readout neuron. $d_{des}$ and $\hat{d}$ denote a potentially desired readout-neuron output and the better (closest) approximation obtained, respectively. $\|d_{des} - \hat{d}\|_2$ is the $l^2$ norm of the residual vector, which is the minimal distance from $d_{des}$ to the conical hull; therefore, this residual vector is normal to the hull face ($u_1$). The squared $l^2$ norm of the residual vector is integrated over this square to evaluate the input-representation error (Ir(C)) corresponding to C. Cone(C) $\bigcap$ $[0, 1]^2$ is the area of the square (hatched with dark blue lines) over which the $l^2$ norm is 0 (the readout neuron can obtain these desired output values). Consequently, the squared $l^2$ norm only has to be integrated over the region $[0, 1]^2 - Cone(C)$ (gray area).

**FIGURE 4 |** Evaluation of the representation error for a 3-state-and-3-input-neuron network. $u_1$, $u_2$, and $u_3$ represent the column vectors of C. The conical hull of these vectors is represented by the cube central empty space. The input-representation error corresponding to this matrix (Ir(C)) is calculated by integrating the squared $l^2$ norm of the residual vector over the cube of size 1 delimited by the thin yellow line ($[0, 1]^3$). The integration over this cube is divided into five regions, corresponding to the red, blue, green, magenta and orange polyhedrons. These regions are determined by the geometry of the cone defined by matrix C. The region corresponding to the empty space does not need to be considered since its integral is 0. The five integrals are summed to obtain Ir(C).

Input matrices including three states can be represented in a three-dimensional space, as in the following matrix C of a 3-input-neuron network:

$$C = \begin{pmatrix} 2 & 3 & 0 \\ 3 & 1 & 0 \\ 1 & 1 & 1 \end{pmatrix}$$

Its geometrical representation being:

The set of all the considered readout-neuron outputs ($[0, 1]^3$) is represented in **Figure 4** by a cube in the three-dimensional space. The conical hull corresponding to this matrix C occupies the central empty space in this cube. If we regard this conical hull as a three-dimensional geometric shape (a three-dimensional pyramid of infinite height without base), it comprises several elements of lower dimensionality: three faces (each one is defined by a couple of vectors $u_k$ and is two-dimensional since a face is contained in a plane), and three edges (each one is defined by a vector $u_k$, that is, a ray, and is one-dimensional). The distance from a point $d_{des}$ to the hull is equivalent to the distance from this point to the closest hull element. Calculating the distance to one of these lower-dimensionality elements (instead of distance to the entire three-dimensional hull) brings the advantage of obtaining a mathematical expression (instead of an algorithm) for this distance (Equation A1). This distance expression allows for the integration over a set of desired points (a region) of $[0, 1]^3$. In order to integrate this distance (squared $l^2$ norm of the residual vector) over the entire cube ($[0, 1]^3$) the cube is partitioned into regions (polyhedrons). Each region contains the points of the cube that are closer to a distinct hull element (i.e., face or edge). Finally, the results of the definite integrations over these regions (or volumes) are summed to obtain Ir(C).

In a three-dimensional space the set of possible desired outputs ($[0, 1]^3$) can be partitioned into at most six regions when the conical hull is defined by 3 rays only (plus the volume occupied by the conical hull). In the case of this matrix $C$, the set of points corresponding to the ray $u_3$ (edge) is empty, so only five integrations are to be calculated.

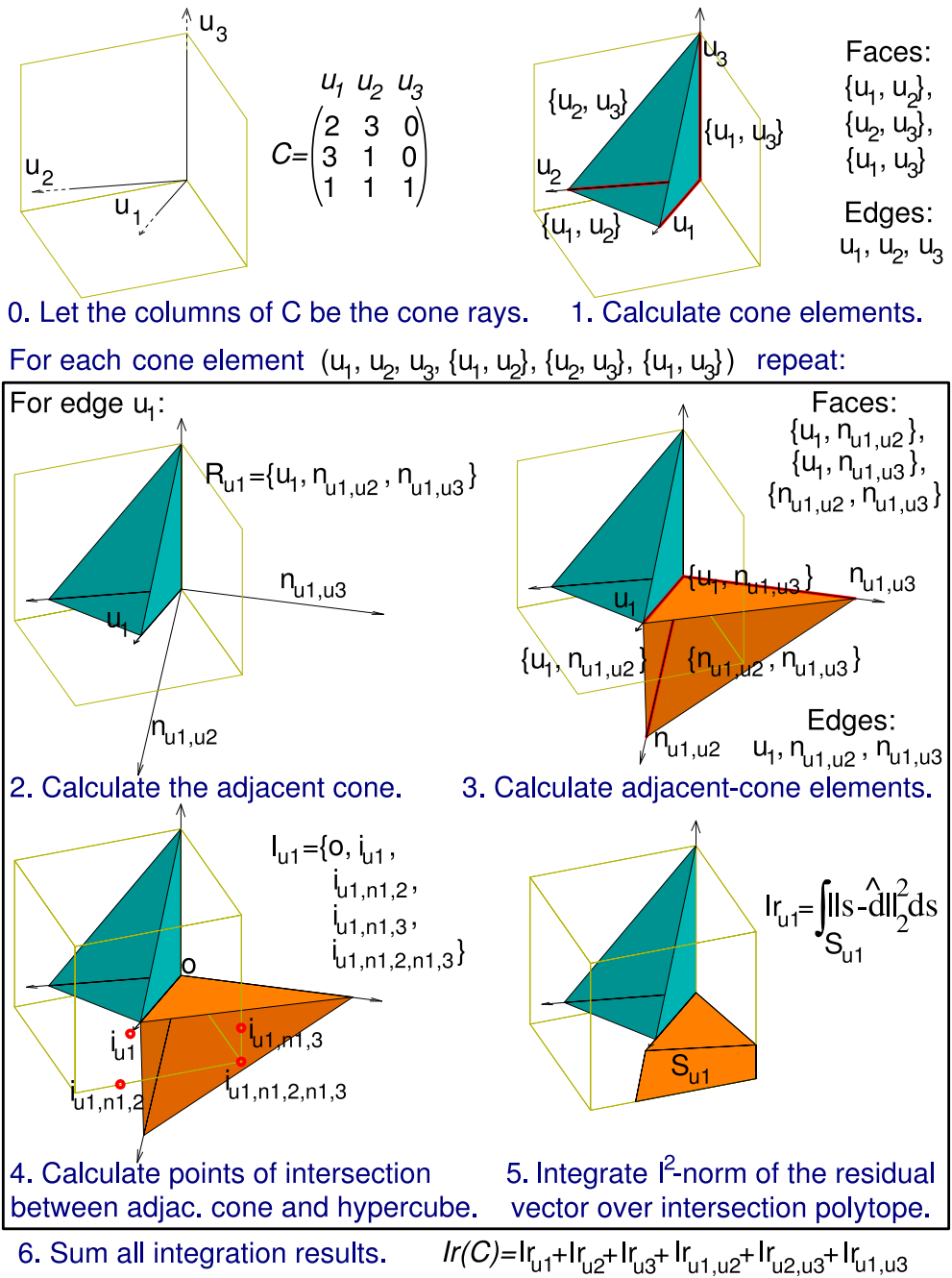

**FIGURE 5 |** Calculation of the evaluation function. The computation of $lr(C)$ can be decomposed into a series of sub-calculations (numbered here from 0 to 5). Text in blue indicates the general calculation procedure and text in black provides the particular calculation results when using the previous 3-by-3 matrix $C$. These sub-calculations are also represented graphically in a three-dimensional space, since the column vectors of matrix $C$ ($u_1$, $u_2$, and $u_3$) have 3 components each. Sub-calculations 2, 3, 4, and 5 must be repeated for every geometrical element of the initial cone (in the case of this matrix $C$ the initial cone comprises 6 elements: 3 faces and 3 edges), but for the sake of brevity we only show these sub-calculations for the edge $u_1$. Therefore, the calculation of only 1 adjacent cone ($R_{u1}$) is showed. $n_{ul,uk}$ denotes an unit vector that is normal to the initial-cone face $\{u_l, u_k\}$, that is, the face defined by rays $u_l$ and $u_k$. This adjacent cone leads to the calculation of its intersection with the cube ($I_{ul}$) and the integration ($Ir_{u1}$) of the squared distance over the corresponding polyhedron ($S_{u1}$). The intersection is decomposed into groups of sub-intersections (in the case of this 3-dimensional space, we have 3 groups, one for each element type of the adjacent cone: edges, faces, and cone inside). $i_{ul}$ denotes the point resulting from the intersection of cone edge $u_l$ and a cube face, $i_{ul,nk,p}$ denotes the intersection point for cone face $\{u_l, n_{k,p}\}$ and a cube edge, and $i_{ul,nk,p,ns,t}$ denotes the intersection point for cone (inside) $\{u_l, n_{k,p}, n_{s,t}\}$ and a cube vertex.

In a higher dimensional space of $m$-input states, the $m$-dimensional cone defined by the conical hull comprises elements from $m$-1 dimensions (facets) to 1 dimension (edges), i.e., $m$-1 dimensions (facets), $m$-2 dimensions (ridges),..., 3 dimensions (cells), 2 dimensions (faces), and 1 dimension (edges). Each cone element is associated with the set of points (region) of the $m$-dimensional hypercube ($[0, 1]^m$) that are closer to this element. Each of these regions constitutes an $m$-dimensional geometric object called polytope, which is the $m$-dimensional version of the polyhedron. Therefore, we must identify all the cone elements to calculate all the hypercube regions over which the squared $l^2$ norm of the residual vector is integrated.

The total value of the squared-distance integration over these regions ($Ir(C)$) is minimal (0) when the conical hull covers the entire hypercube (cube I of **Figure 7**), and maximal when all the elements of the matrix $C$ are zero (cube A of **Figure 7**). A normalized version of $Ir(C)$ can be obtained by dividing the value of $Ir(C_{mxn})$ by this maximal value ($Ir(0_{mx1})$). This maximal value can be calculated by integrating the distance from each point in $[0, 1]^m$ to the only output achieved by this network (0). The results of this integration ($m/3$) is included in this normalized version of $Ir(C)$ as follows:

$$IrN(C) = \frac{Ir(C)}{Ir(0_{m,1})} = \frac{Ir(C)}{\int_S \|s - 0_{mx1}\|_2^2 \, ds} = \frac{Ir(C)}{m/3} \qquad (7)$$

where $m$ refers to the number of rows of $C$, $0_{mx1}$ denotes the zero column vector of size $m$ (included for clarity) and $S$ is the $[0, 1]^m$ space. Thus, $IrN(C)$ provides values in the interval $[0, 1]$. A value of 0 indicates the best input-state representation whereas a value of 1 indicates the worst.

## Computation of the Evaluation Function

The computation of $Ir(C)$ can be decomposed into the following sub-calculations:

0) The column vectors of matrix $C$ are used to define the initial conical hull.
1) All the geometrical elements of the resulting cone (facets, ridges,..., faces and edges) are calculated.

   For each of these elements:

2) The rays of the cone adjacent to the current element are calculated.
3) All the geometrical elements of this adjacent cone are calculated.
4) The vertices of the intersection (region) between the adjacent code and the hypercube $[0, 1]^m$ are calculated.
5) The squared $l^2$ norm of the residual vector (squared Euclidean distance between region points and the initial conical hull) is integrated over the region.

   Finally:

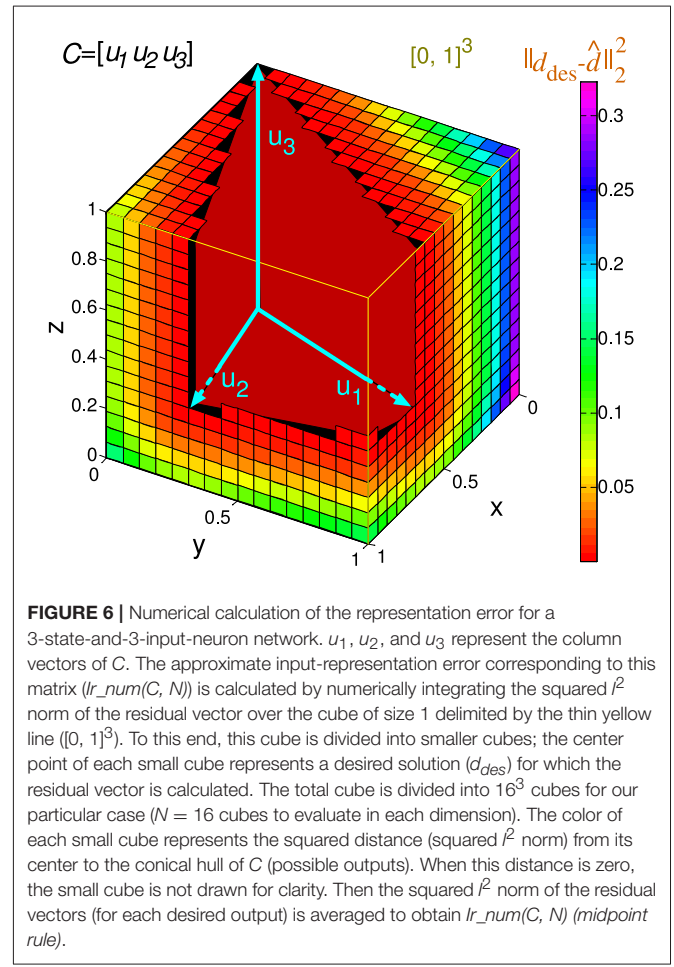

**FIGURE 6 |** Numerical calculation of the representation error for a 3-state-and-3-input-neuron network. $u_1$, $u_2$, and $u_3$ represent the column vectors of $C$. The approximate input-representation error corresponding to this matrix ($Ir\_num(C, N)$) is calculated by numerically integrating the squared $l^2$ norm of the residual vector over the cube of size 1 delimited by the thin yellow line ($[0, 1]^3$). To this end, this cube is divided into smaller cubes; the center point of each small cube represents a desired solution ($d_{des}$) for which the residual vector is calculated. The total cube is divided into $16^3$ cubes for our particular case ($N = 16$ cubes to evaluate in each dimension). The color of each small cube represents the squared distance (squared $l^2$ norm) from its center to the conical hull of $C$ (possible outputs). When this distance is zero, the small cube is not drawn for clarity. Then the squared $l^2$ norm of the residual vectors (for each desired output) is averaged to obtain $Ir\_num(C, N)$ (midpoint rule).

6) The integration results corresponding to all the calculated regions are summed to obtain $Ir(C)$.

A detailed description of these calculations and their implementation is provided in **Appendix A** in supplementary material.

We have used the previous matrix $C$ of a 3-input-neuron network to illustrate these calculations (**Figure 5**).

## Numerical Approximation of the Evaluation Function

A numerical approximation of $Ir(C)$ (and $IrN(C)$) is obtained and compared to the analytical solution previously formulated for illustrative and validation purposes. This approximation numerically integrates the squared $l^2$ norm of the residual vector over the hypercube $[0, 1]^m$, that is, it calculates the average error performed by the network for a large set of desired outputs. We implemented this integration through uniform distance-function sampling [rectangle method with midpoint approximation or midpoint rule (Davis and Rabinowitz, 2007)] over $[0, 1]^m$:

$$Ir\_num(C, N) = \frac{1}{N^m} \sum_{n_1=1}^{N} \sum_{n_2=1}^{N} \cdots \sum_{n_m=1}^{N} \left\| \frac{(n_1, n_2, ..., n_m) - 1/2}{N} \right.$$

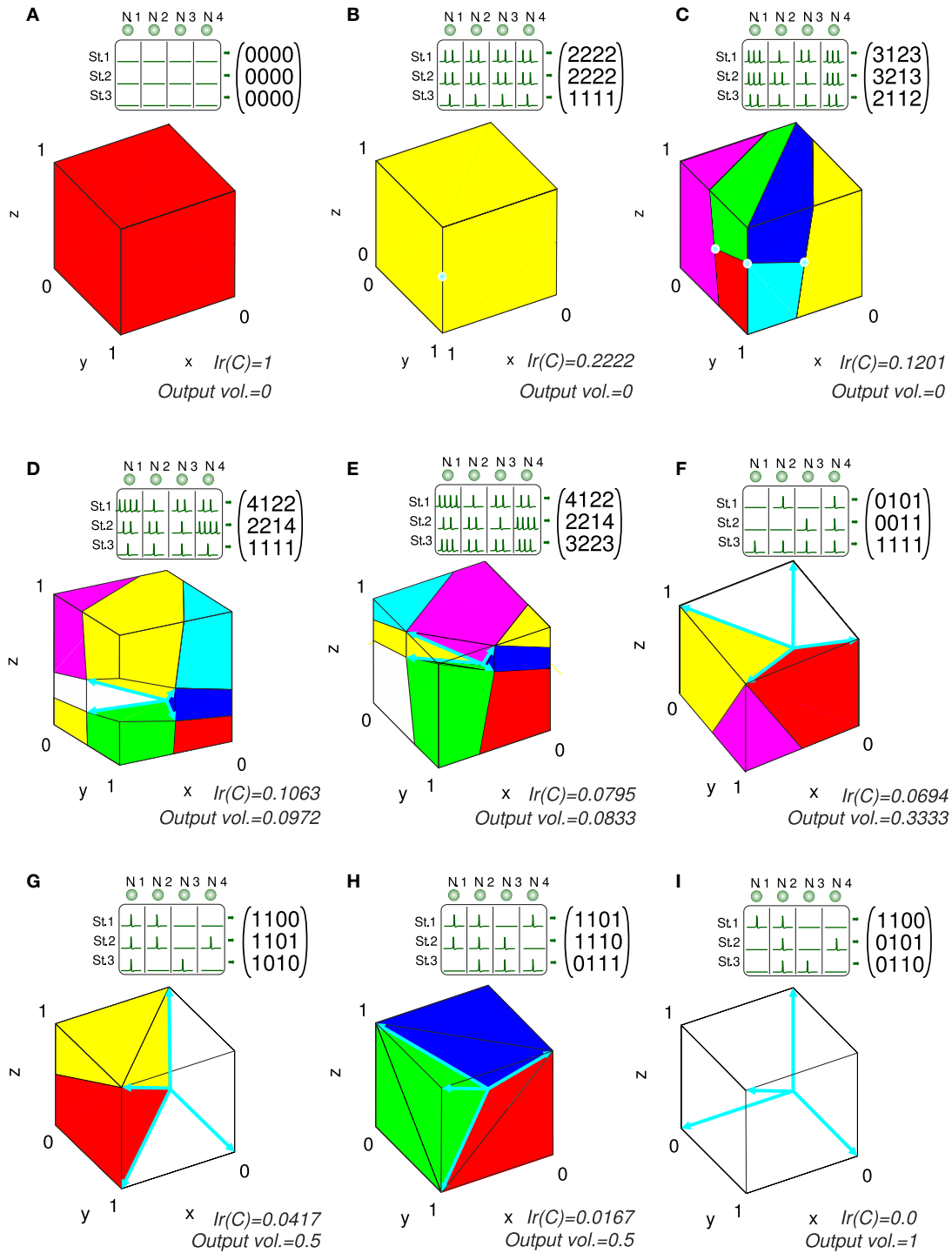

**FIGURE 7 |** Analytical calculation and graphic representation of the input-representation error ($Ir(C)$) for a 3-state and 4-input-neuron network. Nine input matrices $C$ are evaluated, from worst **(A)** to best-case representation **(I)** by way of intermediate cases **(B-H)**. In each case, the input matrix $C$ and its corresponding activity representation (spikes) are shown. The four input neurons (N. 1,..., N. 4) are represented in a three-dimensional space (3 states: St. 1,..., St. 3) by four vectors (colored in cyan), constituting the columns of the input matrix. Redundant vectors (neurons), those that are not extreme rays of the cone, can be identified (they are not edges of the cone). $Ir(C)$ is calculated by integrating the squared $l^2$ norm of the residual vector over the cube [0, 1]³ ($Ir(C)$ and volume values are expressed with a precision of 4 decimal places). The cube is partitioned into regions (colored polyhedrons) over which the squared $l^2$ norm is integrated. The volume of the outputs achieved by the network is also calculated. This volume is represented by the empty space in the cube not occupied by any polyhedron. A larger volume of achieved output values usually results in a better input representation (lower $Ir(C)$), particularly when the volume is centered in the cube.

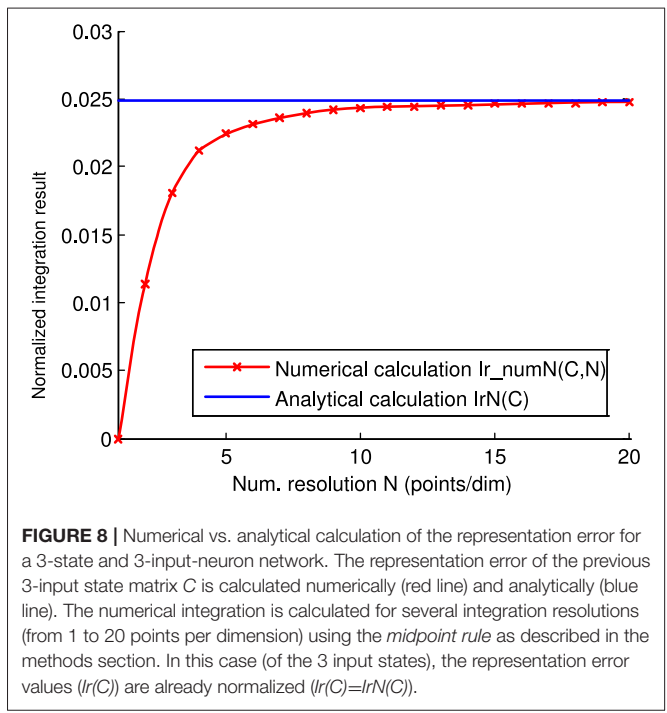

**FIGURE 8 |** Numerical vs. analytical calculation of the representation error for a 3-state and 3-input-neuron network. The representation error of the previous 3-input state matrix $C$ is calculated numerically (red line) and analytically (blue line). The numerical integration is calculated for several integration resolutions (from 1 to 20 points per dimension) using the *midpoint rule* as described in the methods section. In this case (of the 3 input states), the representation error values ($Ir(C)$) are already normalized ($Ir(C)=IrN(C)$).

$$-C\hat{w}_{((n_1,n_2,...,n_m)-1/2)/N}\Big\|_2^2 \qquad (8)$$

where $N$ represents the number of desired points to evaluate in each dimension, $((n_1, n_2,..., n_m)-1/2)/N$ denotes the coordinate vector of the midpoint ($d_{des}$) of each rectangle (row vector of length $m$) and $\hat{w}_s$ is a weight vector that minimizes the readout-neuron error for a desired output $s$. These weights are calculated through an algorithm for the problem of linear least squares with non-negativity constraints for each point $d_{des}$ (Lawson and Hanson, 1974).

Considering the previous matrix $C$ of a 3-input-neuron network with three input states, the numerical calculation of $Ir()$ for $N = 16$ ($Ir\_num(C,16)$) is represented geometrically in **Figure 6**. The hypercube containing the potential readout-neuron outputs is showed in this figure. This hypercube (cube in this case) is divided into small cubes. Each one represents a considered output that has been evaluated. The color of these small cubes codifies the error committed by the readout neuron for each desired output.

The resulting value of this numerical calculation is represented in **Figure 8**. In a similar manner to $Ir(C)$, we can subsequently normalize $Ir\_num(C,N)$ ($Ir\_numN(C,N)$) to obtain values within the [0, 1] interval.

## RESULTS

### Implementation

The presented algorithm was implemented by the application software called AVERPOIN, which is written in MATLAB language. This application can be executed by MATLAB and Octave and is free software provided under the GNU Lesser General Public License version 3 (https://github.com/rrcarrillo/AVERPOIN).

This functional implementation of the presented algorithm is provided as proof of concept. This implementation is not optimized in terms of computational load. Future research to improve the algorithm itself by adopting more efficient methods (De Loera et al., 2012) is advisable. The computational load generated by the integration of the squared $l^2$ norm of the residual vector over the polytopes is expected to be significantly reduced.

### Analytical Solution, Proof of Concept

We present several 3-state-and-4-input-neuron matrix examples to further illustrate the functionality of the algorithm. The software AVERPOIN facilitated the analytical calculation of a floating-point number ($Ir(C)$) that indicated the suitability of each input matrix to generate any output set. The graphical representation helped us to better associate the resulting number, given by the algorithm, and the goodness of each matrix as input-state representation (**Figure 7**). In the case of 3-state representations ($m = 3$), the maximal value of $Ir(C)$ is 1, therefore, $Ir(C)=IrN(C)$. For each matrix, we show the polyhedrons into which the output space was partitioned for the integration process. The input matrices selected as an example cover a range of input representations from worst (A) to best-case scenario (I) showing a gradual decrease in the representation goodness. Worst-case ($Ir(C)=1$) and best-case scenarios ($Ir(C)=0$) were trivial evaluations: however, non-discernible configurations to the naked eye were also sorted and addressed. The matrix columns were represented by cyan vectors (only the intersection points of vectors and cube faces were shown in B and C). Note that the magnitude of these vectors did not affect the representation error, and it is only their direction which determined the corresponding cone ray. This vector representation provided enlightenment about which input neurons (matrix columns) were more informative for the network. In A, B, C, G, H, and I one or more input neurons carried redundant information, and thereby less than four vectors constituted cone extreme rays. For each matrix, we also provided the space volume corresponding to the set of possible outputs generated by the network (*output vol.*). Different input representations may achieve the same number (volume) of readout-neuron outputs (*output vol.* of G and H) but with distant goodness values ($Ir(C)$). This divergence arises from the different locations of this output volume (zero-error outputs defined by the conical hull) within the cube: In H, the zero-error-outputs were mainly confined to a corner of the cube (distant from most of the other outputs), whereas in G, a greater number of non-zero-error outputs were adjacent to the zero-error outputs. These adjacent non-zero-error outputs generate lower errors (the error of a potential output depends on its squared distance to the closest zero-error output) resulting in a lower overall output error. Similarly, the matrix in E achieved a lower volume of zero-error outputs, but these outputs were more centered in the cube, resulting in a better representation and lower $Ir(C)$ than in D (see **Figure 6** for an insight into the effect of the conical hull on the error of adjacent outputs).

## Analytical and Numerical Calculation

We calculated the representation error of the previous 3-state and 3-input-neuron matrix $C$ analytically (**Figure 4**) and numerically (**Figure 6**). The corresponding analytical and numerical-calculation values for different numerical resolutions were calculated and compared (**Figure 8**). The numerical calculation converged to the analytical calculation as the integration resolution was increased, thus lending validity to the effectiveness of both calculations.

## DISCUSSION

Determining the suitability of a particular input activity for a neural circuit is pivotal in neuroscience when studying and modeling networks. We focused on the case of supervised-learning networks with positive inputs. This type of network is frequently employed when modeling circuits such as the cerebellar granular layer and the avian inferior colliculus. We addressed this problem by means of an algorithm able to measure the effectiveness of an input-state activity representation when the network tries to learn a varied set of outputs. This effectiveness measure not only considers the amount of different output sets achieved by the network but also how these output sets are located in the output space in order to exactly determine the goodness of the representation. The algorithm here proposed aims to improve the understanding of brain circuits by establishing a quantifiable association between the nervous-circuit physical structure (i.e., neurobiological network topology substrate) and neural activity. This measurement algorithm can provide insights into information representations in terms of spike trains. It can also suggest changes in the neural circuits to generate more effective activity or to better fit this activity. For example, the input neurons carrying redundant information can be readily identified, and the operation of certain nervous circuits generating an adequate input-state representation (such as the cerebellar granular layer) for the subsequent neurons (e.g., Purkinje cells) can also be readily evaluated in terms of distinguishable states. This evaluation can help to identify which specific neural processing properties and mechanisms (e.g., synaptic plasticity) improve the state representation and therefore are functionally relevant. Moreover, when modeling large-scale neural circuits to investigate their operation, some of their specific information-processing properties are usually not properly considered. Sometimes these specific information-processing properties are not implemented (simple neural and network models) and sometimes, they are implemented (realistic and computationally-expensive models) but the model parameters, although derived from experimental determinations, initially may not be accurate enough to correctly take advantage of these properties from a functional point of view. Thus, combinatorial optimization (such as evolutionary algorithms) could be used to find an optimal value for these parameters, using the presented algorithm as the objective function.

To sum up, the algorithm presented here is, to the best of our knowledge, the first algorithm able to analytically evaluate the suitability (fitness) of an input representation (input activity) for a supervised-learning network with positive inputs to learn any output without actual training (i.e., without requiring intensive network simulations).

## AUTHOR CONTRIBUTIONS

ER conceived the initial idea. RC and NL designed the algorithm, prepared figures and drafted the manuscript. RC and FN implemented the algorithm. NL designed the algorithm tests. All authors reviewed the manuscript and approved the final version.

## FUNDING

This research was supported by the European Union (H2020-MSCA-IF-2014 658479 SpikeControl and HBP-SGA2 H2020-RIA. 785907 (ER)), Spanish MINECO (TIN2016-81041-R partially funded by FEDER and IJCI-2016-27385), Program 7 of the 2015 research plan of the University of Granada (project ID 16), and the GENIL grant PYR-2014-6 from CEI BioTic Granada.

## ACKNOWLEDGMENTS

We are grateful to our colleague Dr. Karl Pauwels for his comments, which have greatly assisted the research.

## REFERENCES

Abbott, L. F., and Regehr, W. G. (2004). Synaptic computation. *Nature* 431, 796–803. doi: 10.1038/nature03010

Albus, J. S. (1971). The theory of cerebellar function. *Math. Biosci.* 10, 25–61. doi: 10.1016/0025-5564(71)90051-4

Albus, J. S. (1975). A new approach to manipulator control: The cerebellar model articulation controller (CMAC). *J. Dyn. Syst. Measure. Control* 97, 220–227. doi: 10.1115/1.3426922

Arfken, G. (1985). *Gram-Schmidt Orthogonalization, Mathematical Methods for Physicists, 3rd Edn.* Orlando, FL: Academic Press, 516–520.

Avis, D. (1980). On the extreme rays of the metric cone. *Can. J. Math.* 32, 126–144. doi: 10.4153/CJM-1980-010-0

Barak, O., Rigotti, M., and Fusi, S. (2013). The sparseness of mixed selectivity neurons controls the generalization–discrimination trade-off. *J. Neurosci.* 33, 3844–3856. doi: 10.1523/JNEUROSCI.2753-12.2013

Barber, C. B., Dobkin, D. P., and Huhdanpaa, H. (1996). The quickhull algorithm for convex hulls. *ACM Trans. Math. Softw.* 22, 469–483. doi: 10.1145/235815.235821

Carr, C. E., Fujita, I., and Konishi, M. (1989). Distribution of GABAergic neurons and terminals in the auditory system of the barn owl. *J. Comp. Neurol.* 286, 190–207. doi: 10.1002/cne.902860205

Carrillo, R. R., Ros, E., Barbour, B., Boucheny, C., and Coenen, O. (2007). Event-driven simulation of neural population synchronization facilitated by electrical coupling. *Biosystems* 87, 275–280. doi: 10.1016/j.biosystems.2006.09.023

Carrillo, R. R., Ros, E., Boucheny, C., and Olivier, J. M. C. (2008a). A real-time spiking cerebellum model for learning robot control. *Biosystems* 94, 18–27. doi: 10.1016/j.biosystems.2008.05.008

Carrillo, R. R., Ros, E., Tolu, S., Nieus, T., and D'Angelo, E. (2008b). Event-driven simulation of cerebellar granule cells. *Biosystems* 94, 10–17. doi: 10.1016/j.biosystems.2008.05.007

Cayco-Gajic, A., Clopath, C., and Silver, R. A. (2017). Sparse synaptic connectivity is required for decorrelation and pattern separation in feedforward networks. *Nat. Commun.* 8:1116. doi: 10.1038/s41467-017-01109-y

Chen, D., and Plemmons, R.J. (2009). "Nonnegativity constraints in numerical analysis," in *The Birth of Numerical Analysis, 1st Edn.*, ed A. Bultheel (Leuven; Singapore: Katholieke Universiteit Leuven; World Scientific), 109–140. doi: 10.1142/9789812836267_0008

Clopath, C., Badura, A., De Zeeuw, C. I., and Brunel, N. (2014). A cerebellar learning model of vestibulo-ocular reflex adaptation in wild-type and mutant mice. *J. Neurosci.* 34, 7203–7215. doi: 10.1523/JNEUROSCI.2791-13.2014

Cooke, S. F., and Bliss, T. V. P. (2006). Plasticity in the human central nervous system. *Brain* 129, 1659–1673. doi: 10.1093/brain/awl082

Dan, Y., and Poo, M. M. (2004). Spike timing-dependent plasticity of neural circuits. *Neuron* 44, 23–30. doi: 10.1016/j.neuron.2004.09.007

D'Angelo, E., Koekkoek, S. K. E., Lombardo, P., Solinas, S., Ros, E., Garrido, J., et al. (2009). Timing in the cerebellum: oscillations and resonance in the granular layer. *Neuroscience* 162, 805–815. doi: 10.1016/j.neuroscience.2009.01.048

Davis, P. J., and Rabinowitz, P. (2007). *Methods of Numerical Integration, 2nd Edn.* New York, NY: Courier Corporation; Dover Publications Inc.

De Loera, J. A., Dutra, B., Koeppe, M., Moreinis, S., Pinto, G., and Wu, J. (2012). Software for exact integration of polynomials over polyhedra. *ACM Comm. Comp. Algebra* 45, 169–172. doi: 10.1145/2110170.2110175

Dean, P., and Porrill, J. (2014). Decorrelation learning in the cerebellum: computational analysis and experimental questions. *Prog. Brain Res.* 210, 157–192. doi: 10.1016/B978-0-444-63356-9.00007-8

Doya, K. (1999). What are the computations of the cerebellum, the basal ganglia and the cerebral cortex? *Neural Netw.* 12, 961–974. doi: 10.1016/S0893-6080(99)00046-5

Doya, K. (2000). Complementary roles of basal ganglia and cerebellum in learning and motor control. *Curr. Opin. Neurobiol.* 10, 732–739. doi: 10.1016/S0959-4388(00)00153-7

Elgersma, Y., and Silva, A. J. (1999). Molecular mechanisms of synaptic plasticity and memory. *Curr. Opin. Neurobiol.* 9, 209–213. doi: 10.1016/S0959-4388(99)80029-4

Feng, J. (ed.). (2003). *Computational Neuroscience: A Comprehensive Approach.* Boca Raton, FL; London, UK: Chapman & Hall/CRC Press. doi: 10.1201/9780203494462

Friedel, P., and van Hemmen, J. L. (2008). Inhibition, not excitation, is the key to multimodal sensory integration. *Biol. Cybern.* 98, 597–618. doi: 10.1007/s00422-008-0236-y

Fujita, M. (1982). Adaptive filter model of the cerebellum. *Biol. Cybern.* 45, 195–206. doi: 10.1007/BF00336192

Gandolfi, D., Lombardo, P., Mapelli, J., Solinas, S., and D'Angelo, E. (2013). Theta-frequency resonance at the cerebellar input stage improves spike timing on the millisecond time-scale. *Front. Neural Circ.* 7:64. doi: 10.3389/fncir.2013.00064

Garrido, J. A., Luque, N. R., D'Angelo, E., and Ros, E. (2013). Distributed cerebellar plasticity implements adaptable gain control in a manipulation task: a closed-loop robotic simulation. *Front. Neural Circ.* 7:159. doi: 10.3389/fncir.2013.00159

Garrido, J. A., Luque, N. R., Tolu, S., and D'Angelo, E. (2016). Oscillation-driven spike-timing dependent plasticity allows multiple overlapping pattern recognition in inhibitory interneuron networks. *Int. J. Neural Syst.* 26:1650020. doi: 10.1142/S0129065716500209

Gerstner, W., and Kistler, W. M. (2002). *Spiking Neuron Models: Single Neurons, Populations, Plasticity.* Cambridge, UK: Cambridge University Press. doi: 10.1017/CBO9780511815706

Häusser, M., and Clark, B. A. (1997). Tonic synaptic inhibition modulates neuronal output pattern and spatiotemporal synaptic integration. *Neuron* 19, 665–678. doi: 10.1016/S0896-6273(00)80379-7

Haykin, S. (1999). *Neural Networks: A Comprehensive Foundation, 2nd Edn.* Upper Saddle River: Prentice-Hall.

Hiriart-Urruty, J. B., and Lemaréchal, C. (1993). *Convex Analysis and Minimization Algorithms I: Fundamentals, Vol. 305.* Berlin: Springer Science & Business Media.

Jeter, M. (1986). *Mathematical Programming: An Introduction to Optimization, Vol. 102 of Pure and Applied Mathematics.* Boca Raton, FL; London, UK: CRC Press.

Kardar, M., and Zee, A. (2002). Information optimization in coupled audio-visual cortical maps. *Proc. Natl. Acad Sci.U.S.A.* 99, 15894–15897. doi: 10.1073/pnas.252472699

Kawato, M., and Gomi, H. (1992). The cerebellum and VOR/OKR learning models. *Trends Neurosci.* 15, 445–453. doi: 10.1016/0166-2236(92)90008-V

Kawato, M., Kuroda, S., and Schweighofer, N. (2011). Cerebellar supervised learning revisited: biophysical modeling and degrees-of-freedom control. *Curr. Opin. Neurobiol.* 21, 791–800. doi: 10.1016/j.conb.2011.05.014

Kempter, R., Gerstner, W., and van Hemmen, J. L. (1999). Hebbian learning and spiking neurons. *Phys. Rev. E* 59, 4498–4514. doi: 10.1103/PhysRevE.59.4498

Khosravifard, M., Esmaeili, M., and Saidi, H. (2009). Extension of the Lasserre–Avrachenkov theorem on the integral of multilinear forms over simplices. *Appl. Math. Comput.* 212, 94–99. doi: 10.1016/j.amc.2009.02.005

Knudsen, E. I. (1994). Supervised learning in the brain. *J. Neurosci.* 14, 3985–3997. doi: 10.1523/JNEUROSCI.14-07-03985.1994

Knudsen, E. I. (2002). Instructed learning in the auditory localization pathway of the barn owl. *Nature* 417, 322–328. doi: 10.1038/417322a

Korbo, L., Andersen, B. B., Ladefoged, O., and Møller, A. (1993). Total numbers of various cell types in rat cerebellar cortex estimated using an unbiased stereological method. *Brain Res.* 609, 262–268. doi: 10.1016/0006-8993(93)90881-M

Lasserre, J. B., and Avrachenkov, K. E. (2001). The Multi-Dimensional Version of $\int_a^b x^p dx$. *Amer. Math. Mon.* 108, 151–154. doi: 10.1080/00029890.2001.11919735

Lawson, C. L., and Hanson, R. J. (1974). *Solving Least Squares Problems, Vol. 161.* Englewood Cliffs, NJ: Prentice-hall.

Lay, D. C., Lay, S. R., and McDonald, J. J. (2014). *Linear Algebra and Its Applications, 5th Edn.* Carmel, IN: Pearson Education.

Luque, N. R., Garrido, J. A., Carrillo, R. R., Coenen, O. J. D., and Ros, E. (2011a). Cerebellar input configuration toward object model abstraction in manipulation tasks. *IEEE Trans. Neural Netw.* 22, 1321–1328. doi: 10.1109/TNN.2011.2156809

Luque, N. R., Garrido, J. A., Carrillo, R. R., Olivier, J. M. D. C., and Ros, E. (2011b). Cerebellarlike corrective model inference engine for manipulation tasks. *IEEE Trans. Syst. Man Cybern. B* 41, 1299–1312. doi: 10.1109/TSMCB.2011.2138693

Luque, N. R., Garrido, J. A., Ralli, J., Laredo, J. J., and Ros, E. (2012). From sensors to spikes: evolving receptive fields to enhance sensorimotor information in a robot-arm. *Int. J. Neural Syst.* 22:1250013. doi: 10.1142/S012906571250013X

Lynch, E. P., and Houghton, C. J. (2015). Parameter estimation of neuron models using *in-vitro* and *in-vivo* electrophysiological data. *Front. Neuroinform.* 9:10. doi: 10.3389/fninf.2015.00010

Markram, H., Lübke, J., Frotscher, M., and Sakmann, B. (1997). Regulation of synaptic efficacy by coincidence of postsynaptic APs and EPSPs. *Science* 275, 213–215. doi: 10.1126/science.275.5297.213

Marr, D. (1969). A theory of cerebellar cortex. *J. Physiol.* 202, 437–470. doi: 10.1113/jphysiol.1969.sp008820

Martínez-Cañada, P., Morillas, C., Pino, B., Ros, E., and Pelayo, F. (2016). A computational framework for realistic retina modeling. *Int. J. Neural Syst.* 26:1650030. doi: 10.1142/S0129065716500301

Masoli, S., Rizza, M. F., Sgritta, M., Van Geit, W., Schürmann, F., and D'Angelo, E. (2017). Single neuron optimization as a basis for accurate biophysical modeling: the case of cerebellar granule cells. *Front. Cell. Neurosci.* 11:71. doi: 10.3389/fncel.2017.00071

McDougal, R. A., Morse, T. M., Carnevale, T., Marenco, L., Wang, R., Migliore, M., et al. (2017). Twenty years of ModelDB and beyond: building essential modeling tools for the future of neuroscience. *J. Comput. Neurosci.* 42, 1–10. doi: 10.1007/s10827-016-0623-7

Miles, F. A., and Eighmy, B. B. (1980). Long-term adaptive changes in primate vestibuloocular reflex. I. Behavioral observations. *J. Neurophysiol.* 43, 1406–1425. doi: 10.1152/jn.1980.43.5.1406

Mortari, D. (1997). n-Dimensional cross product and its application to the matrix eigenanalysis. *J. Guidance Control Dyn.* 20, 509–515. doi: 10.2514/3.60598

Raymond, J. L., and Lisberger, S. G. (1998). Neural learning rules for the vestibulo-ocular reflex. *J. Neurosci.* 18, 9112–9129. doi: 10.1523/JNEUROSCI.18-21-09112.1998

Rucci, M., Tononi, G., and Edelman, G. M. (1997). Registration of neural maps through value-dependent learning: modeling the alignment of auditory and visual maps in the barn owl's optic tectum. *J. Neurosci.* 17, 334–352.

Sawtell, N. B. (2010). Multimodal integration in granule cells as a basis for associative plasticity and sensory prediction in a cerebellum-like circuit. *Neuron* 66, 573–584. doi: 10.1016/j.neuron.2010.04.018

Schneider, P., and Eberly, D. H. (2002). *Geometric Tools for Computer Graphics*. San Francisco, CA: Morgan Kaufmann Publishers.

Schweighofer, N., Doya, K., and Lay, F. (2001). Unsupervised learning of granule cell sparse codes enhances cerebellar adaptive control. *Neuroscience* 103, 35–50. doi: 10.1016/S0306-4522(00)00548-0

Singheiser, M., Gutfreund, Y., and Wagner, H. (2012). The representation of sound localization cues in the barn owl's inferior colliculus. *Front. Neural Circ.* 6:45. doi: 10.3389/fncir.2012.00045

Solinas, S., Nieus, T., and D'Angelo, E. (2010). A realistic large-scale model of the cerebellum granular layer predicts circuit spatio-temporal filtering properties. *Front. Cell. Neurosci.* 4:12. doi: 10.3389/fncel.2010.00012

Song, S., Miller, K. D., and Abbott, L. F. (2000). Competitive Hebbian learning through spike-timing-dependent synaptic pasticity. *Nat. Neurosci.* 3, 919–926. doi: 10.1038/78829

Stitt, I., Galindo-Leon, E., Pieper, F., Hollensteiner, K. J., Engler, G., and Engel, A. K. (2015). Auditory and visual interactions between the superior and inferior colliculi in the ferret. *Eur. J. Neurosci.* 41, 1311–1320. doi: 10.1111/ejn.12847

Takahashi, T. T. (2010). How the owl tracks its prey–II. *J. Exp. Biol.* 213, 3399–3408. doi: 10.1242/jeb.031195

Tyrrell, T., and Willshaw, D. (1992). Cerebellar cortex: its simulation and the relevance of Marr's theory. *Philos. Trans. R. Soc. Lond. B Biol. Sci.* 336, 239–257.

Van Geit, W., Gevaert, M., Chindemi, G., Rössert, C., Courcol, J.-D., Muller, E. B., et al. (2016). BluePyOpt: leveraging open source software and cloud infrastructure to optimise model parameters in neuroscience. *Front. Neuroinform.* 10:17. doi: 10.3389/fninf.2016.00017

Wong, Y. F., and Sideris, A. (1992). Learning convergence in the cerebellar model articulation controller. *IEEE Transac. Neural Netw.* 3, 115–121. doi: 10.1109/72.105424

Yamazaki, T., and Tanaka, S. (2007). The cerebellum as a liquid state machine. *Neural Netw.* 20, 290–297. doi: 10.1016/j.neunet.2007.04.004

**Conflict of Interest Statement:** The authors declare that the research was conducted in the absence of any commercial or financial relationships that could be construed as a potential conflict of interest.

# APPENDIX A

## Algorithm for the Evaluation Function

In order to calculate the value of $Ir(C)$ (and $IrN(C)$), we propose the algorithm described by the following pseudocode:

**Algorithm Ir()**
```
Input: C (m-by-n activity representation
matrix of the m input states)
Output: Ir (integration of the squared
l² norm of the residual vector over the
hypercube [0, 1]ᵐ)
Variables: N, I (list of vectors)
       K, Ks, A, Kp, S (bidimensional
array of vectors)
       E, Es (list of bidimensional arrays
of vectors)
       R (m-by-m matrix)
       Irs (scalar)
       // All vectors are m-dimensional
K = Coni_facets(C) // K[i] is the list of
rays (vectors) which define the i-th cone
facet of the convex conical hull defined by
the column vectors of C (calc. 1.a).
E = Cone_sub-elements(K) // E[i][j] is the
list of rays which define the j-th cone
element of dimension i (for facets i =
m-1, for ridges i = m-2,..., and for rays
i = 1) (calc. 1.b).

Ir = 0
//   we iterate through all cone elements in
E:
For i = 1 to m-1 (number of dimensions-1
of the space)
  For j = 1 to length(E[i]) (number of cone
elements of dimension i in E)
    A = Adjacent_facets(E[i][j]) // A[p] is
the list of m-1 rays which define the p-th
cone facet that is adjacent to the element
E[i][j] (calc. 2.a).
    N = Facet_normals(A) // Each element
of the list N is the vector normal to each
facet included in A (pointing outside the
cone). (calc. 2.b).
    R = [E[i][j] N] // R is a matrix whose
column vectors are the rays which define
the adjacent (simplex) cone, which contains
the points that are closer to the element
E[i][j] than to any other element (calc.
2.c).
    Ks = Coni_facets(R) // (calc. 3.a).
    Es = Cone_sub-elements(Ks) // (calc.
3.b).
    I = Hypercube_intersect(Es) // I is
a list of intersection points of the
hypercube hull and the conical hull of R
(calc. 4).
```

```
    Kp = Polytope_facets(I) // Kp[p] is
the list of points (vertices) which define
the p-th polytope facet of the convex hull
defined by the points of I (except origin)
    S = Triangulate_polytope(Kp) // S[k] is
the list of vertices of the k-th simplex
which composes the Kp polytope (calc. 5.a)
    For k = 1 to length(S) (number of
simplexes in S)
      Irs = Squared_distance_integral(S[k],
E[i][j]) // Irs is the integral of the
squared distance to the element E[i][j]
over the simplex S[k] (calc. 5.b).
      Ir = Ir + Irs // (calc. 5.c and 6).
    End
  End
End
```

As previously stated, the computation of $Ir(C)$ can be broken down into a series of sub-calculations:

0. Definition of an initial conical hull using the column vectors of $C$ (algorithm input) as cone rays.

   The points inside this conical hull represent the output that the readout neuron can generate using $C$ as input.

1. Calculation of all the geometric elements (facets, ridges,. . ., faces and edges) that compose the initial conical hull.

   Each of these elements is defined by a set of rays, and the size of this set depends on the element dimension: an edge is defined by 1 ray, a face is defined by 2 rays, and so on. The calculation of these elements can be divided into 2 steps:

   a. Calculation of the minimal set of facets that define the hull.

   The facets are the sub-elements of higher dimension that compose the hull. They are defined by $m$-1 rays. In a three-dimensional space they are the faces (of 2 rays).

   This facet calculation (Coni facets()) has been implemented through the Quickhull algorithm (Barber et al., 1996). This facet calculation of the conical hull is particularly efficient when the number of input neurons remains less than or equal to the number of input states ($n \leq m$).

   In this step, all the rays that are not extreme are discarded: these rays are not part of the hull boundaries, thus, they are not of the hull facets either. Each ray corresponds to an input neuron; therefore, the neurons that convey redundant information are identified at this step.

   b. Calculation of the lower-dimensionality elements that define the hull from the set of facets.

   These elements (ridges,. . ., cells, faces and edges) are calculated by identifying rays shared between different facets (Cone_sub-elements()).

For each of these calculated cone elements its corresponding region of the hypercube is calculated and the squared distance is integrated over this region. Therefore, sub-calculations (2), (3), (4), and (5) are repeated for each of these elements:

2. Calculation of a new conical hull that is adjacent to the current element.

   This new conical hull constitutes a simplex cone, i.e., it is defined by only $m$ linearly-independent rays (R). This conical hull contains all points in the space that are closer to the current element than to any other element. It is calculated by:

   a. Locating the facets of the initial hull that are adjacent to current element.

      If the current element is a facet, we use it as the adjacent facet.

   b. For each of these adjacent facets, we calculate its perpendicular (normal) ray. These normal vectors are calculated by means of the Laplace expansion of the determinant (Mortari, 1997).

   c. The adjacent hull is finally obtained by combining these normal rays and the rays that define the current element.

   We just need to evaluate the points (potentially desired outputs) in the set (hypercube) $[0, 1]^m$, therefore, we discard the part (desired outputs) of the adjacent hull located outside the hypercube through sub-calculation (3) and (4).

3. Calculation of all the geometric elements (facets, ridges,..., faces and edges) that compose the adjacent cone. These elements are obtained by:

   a. Calculating the facets of the adjacent cone as in step 1.a.

   b. Calculating the lower-dimensionality elements of the adjacent cone from these facets as in step 1.b.

4. Calculation of the intersection between the adjacent conical hull and the hypercube $[0, 1]^m$.

   This intersection results in a polytope that defines one of the regions into which the hypercube is partitioned (see the five polytopes of **Figure 4**). The vertices of this polytope (I) are calculated in parts: For each element Es$[i][j]$ of the adjacent cone (the cone inside, facets, ridges,..., faces and edges) its intersection with the $(m\text{-}i)$-dimensional elements of the hypercube (vertices, edges, faces,..., ridges and facets) is calculated (the intersection of an $i$-dimensional element with an $(m\text{-}i)$-dimensional element in a $m$-dimensional space ordinarily results in a point) (see **Figure 5**). The total set of intersection points define the vertices of the polytope that bound the current integration region. This description of a polytope through a set of vertices is called V-representation.

5. For each obtained region (polytope) the squared $l^2$ norm of the residual vector is integrated.

   This integration is divided into 2 steps:

   a. The polytope is triangulated.

      This triangulation consist in sub-partitioning into simplices (S$[k]$) to facilitate the integral calculation

(Triangulate_polytope()). A simplex is the $m$-dimensional version of a triangle; it is defined by only $m+1$ vertices. This sub-partition is obtained by calculating a list of facets using the Quickhull algorithm, these facets can then be triangulated by applying a fan triangulation (Schneider and Eberly, 2002).

   b. The squared $l^2$ norm of the residual vector, i.e., squared Euclidean distance from a point to its closest cone element (E$[i][j]$), is integrated over each of these simplices (Squared_distance_integral (S$[k]$, E$[i][j]$)).

      This squared distance from point $d$ to the $i$-dimensional element E$[i][j]$ (subspace) in $m$-dimensional space is defined by the following expression:

$$\left\| d - \hat{d} \right\| = d^T d - \sum_{k1\,=\,1}^{m} \sum_{k2\,=\,1}^{m} \left[ (d^T d)_{k1,\,k2} (B^T B)_{k1,\,k2} \right] \tag{A1}$$

where $B$ denotes an $m$-by-$i$ matrix whose column vectors form an orthonormal basis (Lay et al., 2014). This basis spans the minimal vector subspace that contains the element E$[i][j]$. The columns of $B$ have been obtained by applying the Gram–Schmidt process to the rays of E$[i][j]$ (Arfken, 1985). (M)$_{k1,\,k2}$ refers to the element in the $k1$-th row and $k2$-th column of matrix M. M$^T$ denotes the transpose of M. This expression corresponds to a 2-homogeneous polynomial (polynomial whose nonzero-terms have degree two). The integral of this polynomial over the corresponding simplex (S$[k]$) is exactly calculated by applying Lasserre and Avrachenkov's formula (Lasserre and Avrachenkov, 2001):

$$\int_{S_k} \left\| d - \hat{d} \right\|_2^2 ds = \frac{vol(S_k)}{\binom{m+2}{2}} \sum_{k1\,=\,1}^{m} \sum_{l2\,=\,1}^{m} \left[ S_{k,\,l1}^T \, S_{k,\,l2} \right.$$
$$+ \frac{1}{2} \sum_{k1\,=\,1}^{m} \sum_{k2\,=\,1}^{m} \left[ \left( S_{k,\,l1} S_{k,\,l2}^T \right. \right.$$
$$\left. \left. \left. + S_{k,\,l2} \, S_{k,\,l1}^T \right)_{kl,\,k2} \left( B^T B \right)_{k1,\,k2} \right] \right] \tag{A2}$$

where Equation A2 represents the integration of Equation A1 over the S$[k]$ simplex. vol(S$_k$) stands for the volume of the S$[k]$ simplex, which can be calculated through the absolute value of the determinant of S$[k]$ divided by $m!$, and $S_{k,l}$ denotes a column vector which corresponds to the $l$-th vertex of the $k$-th simplex. See Khosravifard et al. (2009) for other integration examples.

   c. The results of these integrations are summed to obtain the integration over the current region.

6. The final $Ir(C)$ value is calculated by summing the integral value for every region.